\documentclass[a4paper]{article}
\usepackage{subcaption}
\usepackage{cite}
\usepackage{SLTU2018}
\usepackage{caption}
\usepackage{url}

\title{Code-Switching Detection with Data-Augmented Acoustic and Language Models}
\name{Emre Y\i lmaz$^{1,2}$, Henk van den Heuvel$^1$ and David A. van Leeuwen$^3$\thanks{This research is funded by the NWO Project 314-99-119 (Frisian Audio Mining Enterprise).}}
\address{
  $^1$ CLS/CLST, Radboud University, Nijmegen, Netherlands \\
  $^2$ Dept. of Electrical and Computer Engineering, National University of Singapore, Singapore \\
  $^3$ ICIS, Radboud University, Nijmegen, Netherlands}
\email{eleey@nus.edu.sg, \{h.vandenheuvel, d.vanleeuwen\}@let.ru.nl}

\begin{document}

\maketitle
\begin{abstract}
In this paper, we investigate the code-switching detection performance of a code-switching (CS) automatic speech recognition (ASR) system with data-augmented acoustic and language models. We focus on the recognition of Frisian-Dutch radio broadcasts where one of the mixed languages, namely Frisian, is under-resourced. Recently, we have explored how the acoustic modeling (AM) can benefit from monolingual speech data belonging to the high-resourced mixed language. For this purpose, we have trained state-of-the-art AMs on a significantly increased amount of CS speech by applying automatic transcription and monolingual Dutch speech. Moreover, we have improved the language model (LM) by creating CS text in various ways including text generation using recurrent LMs trained on existing CS text. Motivated by the significantly improved CS ASR performance, we delve into the CS detection performance of the same ASR system in this work by reporting CS detection accuracies together with a detailed detection error analysis.
\end{abstract}
\vspace{-0.1cm}
\section{Introduction}
\label{sec:intro}
Detecting the language switches in speech with code-switching (CS) is a relevant application while designing automatic speech recognition (ASR) systems for multilingual societies. Impact of CS and other kinds of language switches on the speech-to-text systems have recently received research interest, resulting in several robust acoustic modeling~\cite{stemmer2001,lyu2006,vu2012,modipa2013,lyudovyk2014,wu2014,yilmaz2016_2,weiner2012,lyu2013,yeong2014,mabokela2014} and language modeling~\cite{li2012,adel2013,adel2014,zeng2017,hamed2017,westhuizen2017} approaches for CS speech. Our research in the FAME! project focuses on developing an all-in-one CS ASR system using a Frisian-Dutch bilingual acoustic and language model that allows language switches~\cite{yilmaz2016_2,yilmaz2016_4}. 

One major challenge when building a CS ASR system is the lack of training speech and text data to train reliable acoustic and language models that can accurately recognize the uttered word and its language. The latter is relevant in language pairs such as Frisian and Dutch with orthographic similarity and shared vocabulary. For this purpose, we have proposed automatic transcription strategies developed for CS speech to increase the amount of training speech data in previous work~\cite{yilmaz2017_2, yilmaz2018}. Thanks to the increased CS training data \cite{yilmaz2018}, we reported improvements in the ASR and CS detection accuracy using fully connected deep neural networks (DNN). Later in \cite{yilmaz2018_2}, we have investigated how to combine the out-of-domain speech data from the high-resourced mixed language with this increased amount of CS speech data for acoustic training of a more recently proposed neural network architecture consisting of time-delay and recurrent layers~\cite{peddinti2017}.

For the language modeling, we have thus far incorporated a standard bilingual language model which is mostly trained on a combination of monolingual text from Frisian and Dutch. The only component in the training text with CS is the transcriptions of the FAME! Corpus~\cite{yilmaz2016} training speech data. To enrich this baseline LM, in~\cite{yilmaz2018_2}, we generate CS text either by training long short-term memory (LSTM) language models on the very small amount of CS text extracted from the transcriptions of the training speech data and synthesize much larger amounts of CS text using these models~\cite{sutskever2011} or by translating Dutch text extracted from the transcriptions of a large Dutch speech corpora. The latter provides CS text as some Dutch words, such as proper nouns (most commonly person, place and institution names) which are a prominent source of code-switching, are not translated to Frisian while the neighboring words are. We further include the transcriptions provided by the best performing automatic transcription strategies described in~\cite{yilmaz2018}.

In this work, we investigate the code-switching detection performance of different ASR systems using these augmented acoustic and language models by presenting the detection performance. Moreover, we compare the frequency of the hypothesized language switches, duration distribution of the monolingual segments, accuracy of the assigned language tags and the most common erroneously recognized word pairs with different language tags to provide further insight into the quality of the hypothesized language switches.
\vspace{-0.1cm}
\section{Frisian-Dutch Radio Broadcast Database}
\label{sec:database}
The bilingual FAME! speech database, which has been collected in the scope of the \textit{Frisian Audio Mining Enterprise} project, contains radio broadcasts in Frisian and Dutch. The Frisian language shows many parallels with Old English. However, nowadays it is under growing influence of Dutch due to long lasting and intense language contact. Almost all speakers of Frisian are at least bilingual, since Dutch is the main language used in education in Frysl\^{a}n.

The FAME! project aims to build a spoken document retrieval system operating on the bilingual archive of the regional public broadcaster Omrop Frysl\^{a}n (Frisian Broadcast Organization). This bilingual data contains Frisian-only and Dutch-only utterances as well as mixed utterances with inter-sentential, intra-sentential and intra-word CS~\cite{myers1989}. To design an ASR system that can handle the language switches, a representative subset of recordings has been extracted from this radio broadcast archive. These recordings include language switching cases and speaker diversity, and have a large time span (1966--2015). For further details, we refer the reader to~\cite{yilmaz2016}.
\vspace{-0.1cm}
\section{Data Augmentation}
\label{sec:aug}
\vspace{-0.1cm}
\subsection{Acoustic Modeling}
\label{ssec:am}
\vspace{-0.1cm}
In previous work, we described several automatic annotation approaches to enable using of a large amount of raw bilingual broadcast data for acoustic model training in a semi-supervised setting. For this purpose, we performed various tasks such as speaker diarization, language and speaker recognition and LM rescoring on raw broadcast data for automatic speaker and language tagging~\cite{yilmaz2017_2, yilmaz2018} and later used this data for acoustic training together with the manually annotated (reference) data. These approaches improved the recognition performance on the CS due to the significant increase in the available CS training data.

Our later efforts focus on the possible improvements in the acoustic modeling that can be obtained using other datasets with a much larger amount of monolingual speech data from the high-resourced mixed language, which is Dutch in our scenario. Previously, adding even a small portion of this Dutch data resulted in severe recognition accuracy loss in the low-resourced mixed language. After the significant increase in the CS training speech data, we explore to what extent one can benefit from the greater availability of resources for Dutch. Given that the Dutch spoken in the target CS context is characterized by the West Frisian accent~\cite{bezooijen1999}, we further include speech data from a language variety of Dutch, namely Flemish, to investigate its contribution towards the accent-robustness of the final acoustic model.
\vspace{-0.3cm}
\subsection{Language Modeling}
\label{ssec:lm}
\vspace{-0.1cm}
Language modeling in a CS scenario mainly suffers from lack of adequate training material, as CS rarely occurs (mostly in an informal context such personal messages and tweets) in written resources. Therefore, finding enough training material for a language model that can model word sequences with CS is very challenging. Our main source for CS text was the transcriptions of the training speech data which comprises a very small proportion of the bilingual training text compared to the monolingual Frisian and monolingual Dutch text.

The CS language model (LM) used in this work is enriched by generating code-switching text in three ways: (1) generating text using recurrent LMs trained on the transcriptions of the training CS speech data, (2) adding the transcriptions of the automatically transcribed CS speech data and (3) translating Dutch text extracted from the transcriptions of a large Dutch speech corpora. The perplexity improvements provided by each component is presented in~\cite{yilmaz2018_2}.
\vspace{-0.2cm}
\section{Experimental Setup}
\label{sec:exps}
\subsection{Databases}
\label{ssec:data}
\begin{table}
\centering
\addtolength{\tabcolsep}{-4.2pt}
\caption{Acoustic data composition of different training setups used in the recognition experiments (in hours)}
\vspace{-0.25cm}
\begin{tabular}{| l | c | c | c | c | c |} 
\hline
Training data & Annot. & Frisian & Dutch & Flemish & Total  \\
\hline\hline
(1) FAME!~\cite{yilmaz2016} & Manual &   8.5   &   3.0  &    -    & 11.5  \\
\hline
(2) Frisian Broad.~\cite{yilmaz2018} & Auto. &   \multicolumn{2}{c|}{125.5}  & - & 125.5  \\
\hline
(3) CGN-NL~\cite{cgn} & Manual & - & 442.5 & - & 442.5 \\
\hline
(4) CGN-VL~\cite{cgn} & Manual & - & - & 307.5 & 307.5 \\
\hline
\end{tabular}
\vspace{-0.5cm}
\label{tab:data}
\end{table}
\vspace{-0.1cm}
\subsubsection{Spoken Material}
\vspace{-0.1cm}
Details of all training data used for the experiments are presented in Table~\ref{tab:data}. The training data of the FAME! speech corpus comprises 8.5 hours and 3 hours of speech from Frisian and Dutch speakers respectively. The development and test sets consist of 1 hour of speech from Frisian speakers and 20 minutes of speech from Dutch speakers each. All speech data has a sampling frequency of 16 kHz.

The raw radio broadcast data extracted from the same archive as the FAME! speech corpus consists of 256.8 hours of audio, including 159.5 hours of speech based on the SAD~\cite{graciarena2016} output. The amount of total raw speech data extracted from the target broadcast archive after removing very short segments is 125.5 hours. We refer to this automatically annotated data as the `Frisian Broadcast' data.

Monolingual Dutch speech data comprises the complete Dutch and Flemish components of the Spoken Dutch Corpus (CGN)~\cite{cgn} that contains diverse speech material including conversations, interviews, lectures, debates, read speech and broadcast news. This corpus contains 442.5 and 307.5 hours of Dutch and Flemish data respectively.
\vspace{-0.2cm}
\subsubsection{Written Material}
\vspace{-0.1cm}
The baseline language models are trained on a bilingual text corpus containing 37M Frisian and 8.8M Dutch words. Almost all Frisian text is extracted from monolingual resources such as Frisian novels, news articles, Wikipedia articles. The Dutch text is extracted from the transcriptions of the CGN speech corpus which has been found to be very effective for language model training compared to other text extracted from written sources. The transcriptions of the FAME! training data is the only source of CS text and contains 140k words. 

Using this small amount of CS text, we train LSTM-LM and generate text with 10M, 25M, 50M and 75M words. The translated CS text contains 8.5M words. Finally, we use the automatic transcriptions provided by the best-performing monolingual and bilingual automatic transcription strategy which contains 3M words in total. The details of these strategies are described in~\cite{yilmaz2018}. The final training text corpus after including the generated text has 107M words.
\vspace{-0.2cm}
\subsection{Implementation Details}
\label{ssec:impdet}
\vspace{-0.1cm}
The recognition experiments are performed using the Kaldi ASR toolkit~\cite{kaldi}. We train a conventional context dependent Gaussian mixture model-hidden Markov model (GMM-HMM) system with 40k Gaussians using 39 dimensional mel-frequency cepstral coefficient (MFCC) features including the deltas and delta-deltas to obtain the alignments for training a lattice-free maximum mutual information (LF-MMI)~\cite{povey2016} TDNN-LSTM~\cite{peddinti2017} AM (1 standard, 6 time-delay and 3 LSTM layers). We use 40-dimensional MFCC as features combined with i-vectors for speaker adaptation~\cite{saon2013}. Further details are provided in \cite{yilmaz2018_2}.

The baseline language models are standard bilingual 3-gram with interpolated Kneser-Ney smoothing and an RNN-LM~\cite{mikolov2010} with 400 hidden units used for recognition and lattice rescoring respectively. The RNN-LMs with gated recurrent units (GRU)~\cite{chung2014} and noise contrastive estimation~\cite{chen2015} are trained. A tied LSTM-LM~\cite{sundermeyer2012} with 650 hidden units per layer and 650-dimensional word embeddings is used for CS text generation which is trained for 40 epochs using the example Pytorch implementation available at \url{https://github.com/pytorch/examples}.
\begin{table*}
\centering
\caption{WER (\%) obtained on the development and test set of the FAME! Corpus - Different AM training data is identified with the numbers which are defined in Table~\ref{tab:data}.}
\addtolength{\tabcolsep}{-3.7pt}
\vspace{-0.3cm}
\begin{tabular}{| l | c | c | c || c c c c | c c c c | c |}
\hline
 \multicolumn{4}{|c|}{} & \multicolumn{4}{c|}{Dev.} & \multicolumn{4}{c|}{Test} & Total\\
\hline
 \multicolumn{4}{|c|}{} & fy & nl & fy-nl & all & fy & nl & fy-nl & all & \\
\hline
 \multicolumn{4}{|c|}{\# of Frisian words} & 9190 & 0 & 2381 & 11,571 & 10,753 & 0 & 1798 & 12,551 & 24,122\\
\hline
 \multicolumn{4}{|c|}{\# of Dutch words} & 0 & 4569 & 533 & 5102 & 0 & 3475 & 306 & 3781 & 8883\\
\hline\hline
ASR System  & AM train data & LM train data & Lang. Tag & \multicolumn{4}{c|}{} & \multicolumn{4}{c|}{} & \\
\hline \hline
Baseline ASR & (1) & Orig. & No  & 36.4 & 43.7 & 48.2 & 40.3 & 31.5 & 39.5 & 47.9 & 35.2 & 37.8 \\
\hline
Baseline ASR & (1) & Orig. & Yes & 37.9 & 48.3 & 53.3 & 43.3 & 32.8 & 42.5 & 51.9 & 37.2 & 40.3 \\
\hline \hline
ASR\_AA      & (1+2) & Orig. & No  & 31.1 & 35.2 & 42.4 & 34.1 & 28.6 & 31.8 & 44.0 & 31.2 & 32.7 \\
\hline
ASR\_AA      & (1+2) & Orig. & Yes & 32.9 & 37.8 & 48.2 & 36.8 & 29.5 & 34.5 & 49.4 & 33.1 & 35.0 \\
\hline \hline 
ASR\_AA\_NL  & (1+2+3) & Orig. & No  & 25.8 & 27.4 & 36.9 & 28.1 & 24.8 & 24.6 & 37.6 & 26.4 & 27.2 \\
\hline
ASR\_AA\_NL  & (1+2+3) & Orig. & Yes & 27.4 & 30.7 & 43.1 & 30.9 & 25.8 & 27.8 & 44.1 & 28.5 & 29.7 \\
\hline \hline
ASR\_AA\_NL-VL & (1+2+3+4) & Orig. & No & 26.4 & 26.9 & 36.2 & 28.1 & 24.5 & 23.2 & 38.5 & 26.0 & 27.1 \\
\hline
ASR\_AA\_NL-VL & (1+2+3+4) & Orig. & Yes & 27.7 & 30.5 & 42.7 & 31.0 & 25.5 & 26.1 & 43.5 & 27.8 & 29.4 \\ 
\hline \hline
ASR\_AA\_NL-VL\_CS-LM & (1+2+3+4) & Orig.+Gen. & No & 24.6 & 26.7 & 33.3 & 26.5 & 22.5 & 22.4 & 32.9 & 23.8 & 25.2 \\
\hline
ASR\_AA\_NL-VL\_CS-LM & (1+2+3+4) & Orig.+Gen. & Yes & 25.9 & 30.2 & 38.0 & 29.1 & 23.5 & 26.7 & 38.6 & 26.0 & 27.6 \\
\hline
\end{tabular}
\label{tab:wer}
\vspace{-0.6cm}
\end{table*}

The bilingual lexicon contains 110k Frisian and Dutch words. The number of entries in the lexicon is approximately 160k due to the words with multiple phonetic transcriptions. The phonetic transcriptions of the words that do not appear in the initial lexicons are learned by applying grapheme-to-phoneme (G2P) bootstrapping~\cite{davel2003,maskey2004}. The lexicon learning is carried out only for the words that appear in the training data using the G2P model learned on the corresponding language. We use the Phonetisaurus G2P system~\cite{novak2015} for creating phonetic transcriptions.
\begin{figure}
  \includegraphics[width=3.05in, trim={0cm 0.5cm 2cm 1.75cm}]{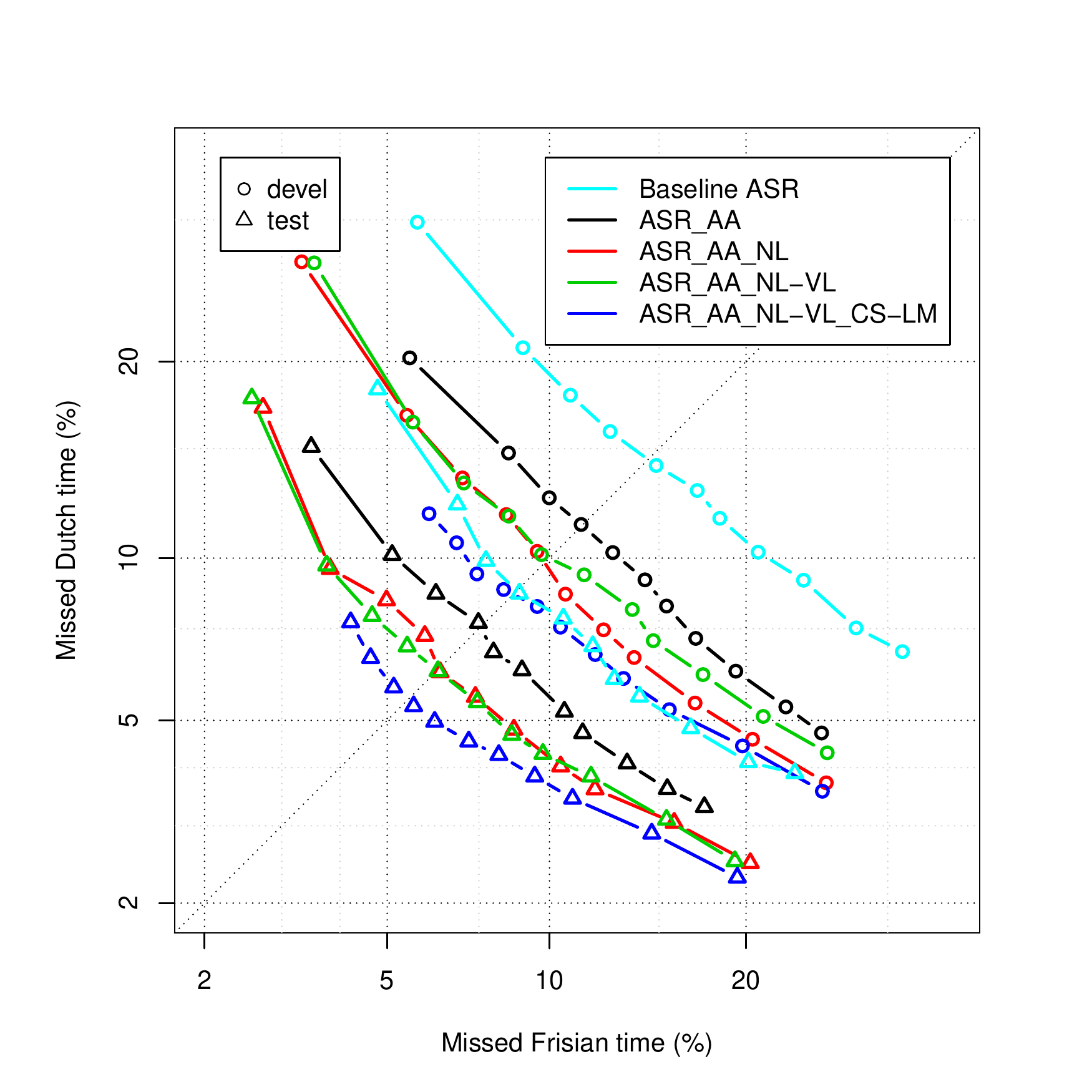}         	
  \vspace{-0.3cm}
  \caption{Code-switching detection performance obtained on the FAME! development and test sets}
  \label{fig:det}
  \vspace{-0.65cm}
\end{figure}
\vspace{-0.2cm}
\subsection{Recognition and CS Detection Experiments}
\label{ssec:exps}
\vspace{-0.1cm}
The baseline system is trained only on the manually annotated data. The other ASR systems incorporate acoustic models trained on the combined data which is automatically transcribed in various ways. These systems are tested on the development and test data of the FAME! speech database and the recognition results are reported separately for Frisian only (fy), Dutch only (nl) and mixed (fy-nl) segments with and without language tags. The results with language tags identify the quality of the assigned language tags for each component. The overall performance (all) is also provided as a performance indicator. The recognition performance of the ASR system is quantified using the Word Error Rate (WER).

After the ASR experiments, we compare the CS detection performance of these recognizers. For this purpose, we use a different LM strategy. We train separate monolingual LMs, and interpolated between them with varying weights, effectively varying the prior for the detected language. For each LM, we generate the ASR output for each utterance. Then, we extract word-level segmentation files for each LM weight. By comparing these alignments with the ground truth word-level alignments, a time-based CS detection accuracy metric is calculated~\cite{yilmaz2017_2}. CS detection accuracy is evaluated by reporting the equal error rates (EER) calculated based on the detection error tradeoff (DET) graph \cite{martin1997} plotted for visualizing the CS detection performance. The presented code-switching detection results indicate how well the recognizer detects the switches and hypothesizes words in the switched language.
\vspace{-0.2cm}
\section{Results and Discussion}
\label{sec:res}
\vspace{-0.1cm}
\subsection{ASR Results}
\label{ssec:asrres}
\vspace{-0.1cm}
The ASR results with and without language tags provided by different ASR systems are presented in Table~\ref{tab:wer}. The baseline ASR trained only on the FAME speech corpus has a total WER of 37.8\% without language tags and 40.3\% with language tags. Applying data augmentation to the AM and LM consistently improves the ASR performance~\cite{yilmaz2018_2}, while the performance gap between the results with and without language tag stays relatively constant between 2\%-2.5\% absolute indicating a relatively constant source of language recognition errors that cannot be eliminated using the augmented AM and LM. 

Furthermore, for all systems, the recognition accuracy on the Dutch only and mixed segments reduces more dramatically due to the language recognition errors compared to the Frisian segments. This suggests the CS LM (yielding the lowest perplexity on the development text) used during the ASR experiments has the tendency to assign Frisian language tags more frequently than Dutch to the phonetically and orthographically similar words. To address these observations, an error analysis on language tag confusions is given in Section~\ref{ssec:dis}.
\vspace{-0.2cm}
\subsection{CS Detection Results}
\label{ssec:csres}
\vspace{-0.1cm}
The CS detection performance provided by the ASR systems on each component of the FAME! development and test data is presented in Figure~\ref{fig:det}. The baseline ASR system provides an EER of 14.4\% (8.7\%) on the development (test) data. Adding automatically annotated in-domain speech and monolingual Dutch speech data (ASR\_AA\_NL) reduces the CS detection accuracy considerably with an EER of 9.8\% (6.3\%). The improved AM yielding a 12.2\% (8.8\%) absolute WER improvement compared to the baseline ASR (cf. Table~\ref{tab:wer}) also improves the quality of the assigned language tags with an absolute EER reduction of 4.6\% (2.4\%).

Comparable to the ASR results, adding monolingual Flemish speech data does not bring any improvement in the CS detection accuracy. Finally, enriching the CS LM using generated textual training data improves the CS detection accuracy by reducing the EER to 8.6\% (5.5\%) which is the best CS detection accuracy among the presented systems.

\begin{figure}[t]
  \includegraphics[width=3.4in, trim={1cm 1.5cm 0cm 0.5cm}]{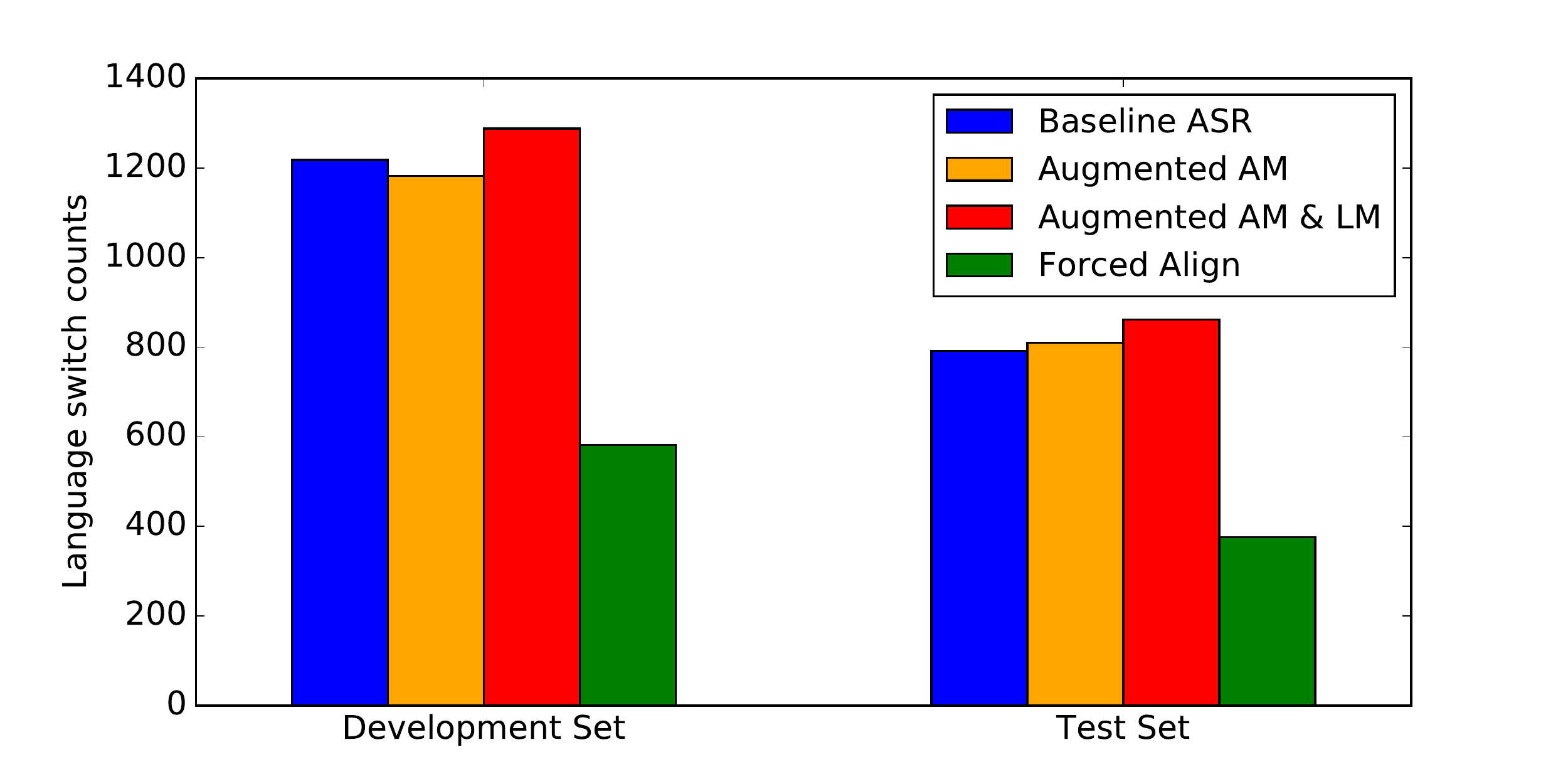}
%  \vspace{-0.3cm}
  \caption{Hypothesized language switch counts}
  \label{fig:lang_swch}
  \vspace{-0.25cm}
\end{figure}

\begin{figure}[t]
  \includegraphics[width=3.4in, trim={1.2cm 1cm 0cm 0.65cm}]{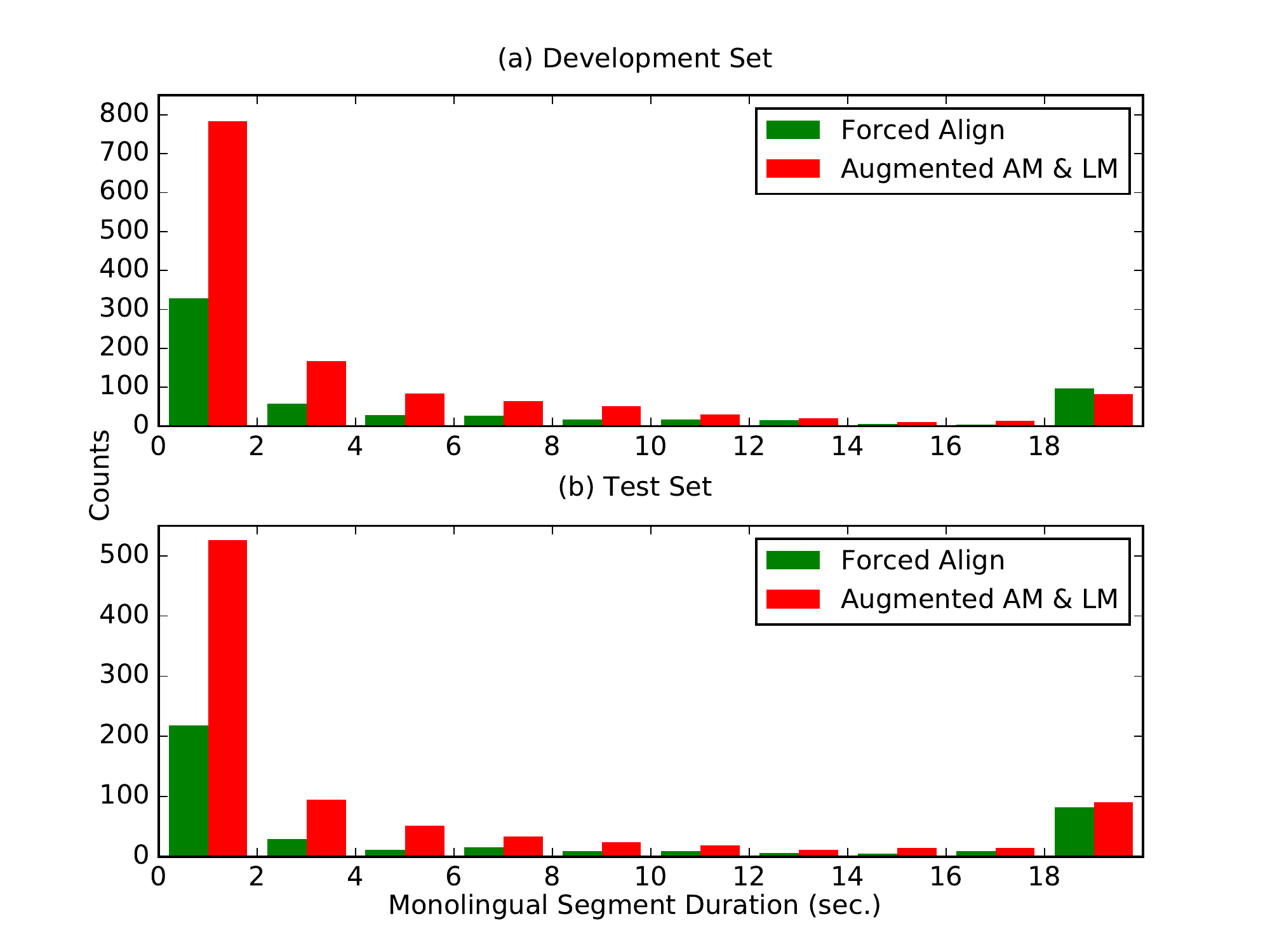}
%   \vspace{-0.3cm}
  \caption{Duration distribution of monolingual segments}
  \label{fig:dur}
  \vspace{-0.72cm}
\end{figure}
\vspace{-0.2cm}
\subsection{Discussion}
\label{ssec:dis}
\vspace{-0.1cm}
After reporting the ASR and CS detection accuracies, we analyze the CS cases hypothesized by different ASR systems. We first present the language switch counts for the baseline ASR, the ASR with the augmented AM (ASR\_AA\_NL-VL) and the ASR with augmented AM and LM (ASR\_AA\_NL-VL\_CS-LM). Later, the histogram plot of the monolingual segment durations provided by the best-performing ASR system in Table~\ref{tab:wer} is compared with the forced alignment output (used as the ground truth). Finally, the most frequent (word-language tag) pair confusions are listed to demonstrate the characteristics of the words that are resulting in CS detection errors.

Figure~\ref{fig:lang_swch} illustrates the language switch counts belonging to different ASR systems. In general, all ASR systems have the tendency to overestimate the number of actual language switches annotated by bilingual Frisian-Dutch speakers. This overestimation in the number of language switches is not noticeable in our duration-based CS detection metric as hypothesizing very short erroneous language switches is less penalized compared to incorrect language tags assigned over longer segments which gives a better indication of the general language recognition capability of the corresponding ASR system. Another observation is the consistent increase in the amount of hypothesized language switches after enriching the CS LM with generated CS textual data containing new CS examples. As expected, the ASR system with augmented AM and LM allows more language switches compared to the other ASR systems on both datasets. 

As a next step, we further compare the distribution of monolingual segments  at the output of the best-performing ASR and forced alignment of the orthographic transcriptions to get a better idea about the overestimation of language switch counts. As it can be seen from Figure~\ref{fig:dur}, the output of the ASR system contains more monolingual segments that are shorter then 10 seconds compared to the forced alignment output. Especially, there is a significant difference in the number of segments that are shorter than 2 seconds. From Figure~\ref{fig:lang_swch} and \ref{fig:dur}, it can be concluded that the best-performing ASR system suffers from false alarms (by hypothesizing very short erroneous language switches) despite providing the best detection performance using a duration-based metric. Investigating smoothing techniques for language tag switches to reduce these false alarm remains as a future work.

In Table~\ref{tab:err}, the most frequently substituted (word-language tag) pairs are listed as the main source of the CS detection errors. From these lists, it can be seen that most common language tag errors are orthographically identical (or similar) short filler words with the same meaning both in Frisian and Dutch. The orthographically identical words, some of which are appearing in this list, induces the 2\%-2.5\% absolute WER difference between the ASR results with and without language tags. However, these errors are less of a concern for the ultimate goal of building a spoken document retrieval system. 

\begin{table}[t]
\caption{Most frequent (word-language tag) confusions of the best-performing ASR system on the development and test data}
\addtolength{\tabcolsep}{-3.8pt}
\vspace{-0.3cm}
\begin{tabular}{| c | c | c || c | c | c |}
\hline
 \multicolumn{3}{|c||}{Dev.} & \multicolumn{3}{c|}{Test} \\
\hline 
Ref. word   & Hyp. word  &  Count &  Ref. word   & Hyp. word  &  Count \\
\hline \hline
en-nl       &   en-fy    &   26   &   en-nl      &    en-fy   &     42 \\
\hline
de-fy       &   de-nl    &   24   &   dat-nl     &   dat-fy   &     33 \\
\hline
dat-nl      &   dat-fy   &   18   &   de-fy      &    de-nl   &     29 \\
\hline
wat-nl      &   wat-fy   &   16   &   is-nl      &    is-fy   &     25 \\
\hline
het-nl      &   it-fy    &   15   &   ja-nl      &    ja-fy   &     24 \\
\hline
dat-fy      &   dat-nl   &   15   &   it-fy      &    't-nl   &     24 \\
\hline
it-fy       &   't-nl    &   14   &   de-nl      &    de-fy   &     23 \\
\hline
is-nl       &   is-fy    &   14   &   het-nl     &    it-fy   &     21 \\
\hline
de-nl       &   de-fy    &   13   &   ja-fy      &    ja-nl   &     20 \\
\hline
in-fy       &  een-nl    &   12   &   dat-fy     &    dat-nl  &     20 \\
\hline
%en-fy       &   en-nl    &   12   &   van-nl     &    fan-fy  &     19 \\
%\hline
%ja-nl       &   ja-fy    &   11   &   wat-nl     &    wat-fy  &     18 \\
%\hline
%fan-fy      &  van-nl    &   11   &   en-fy      &    en-nl   &     17 \\
%\hline
%in-nl       &  yn-fy     &    9   &   een-nl     &    in-fy   &     15 \\
%\hline
%nou-nl      &  no-fy     &    8   &   in-fy      &   een-nl   &     14 \\
%\hline
\end{tabular}
\label{tab:err}
\vspace{-0.65cm}
\end{table}
\vspace{-0.25cm}
\section{Conclusions}
\label{sec:conc}
\vspace{-0.1cm}
This paper explores the CS detection performance of a state-of-the-art CS ASR with data-augmented acoustic and language model. After reporting significant improvements in the ASR accuracy, we explore how well the same ASR system assigns language tags compared to a baseline ASR system using only manually annotated data in the scenario of under-resourced Frisian-Dutch CS speech. The CS detection experiments demonstrate the superior CS detection performance of the data-augmented CS ASR system. Further CS analysis of the ASR output gives insight about the quality of the hypothesized CS cases: (1) the best-performing ASR system hypothesizes more language switches compared to the manual annotations, (2) this overestimation is mostly due to numerous very short erroneous language switches, and (3) the main source of CS detection errors are orthographically identical short filler words which are invisible in the absence of the language tags.
\bibliographystyle{IEEEtran}
\bibliography{refs}
\end{document}